%% file: 0-main.tex
\documentclass{article}

\input{packages}
\input{commands}

\usepackage[textsize=tiny]{todonotes}

\icmltitlerunning{Pragmatic Feature Preferences: Learning Reward-Relevant Preferences from Human Input}

\begin{document}

\twocolumn[
\icmltitle{Pragmatic Feature Preferences: \\Learning Reward-Relevant Preferences from Human Input}



\icmlsetsymbol{equal}{*}

\begin{icmlauthorlist}
\icmlauthor{Andi Peng}{MIT}
\icmlauthor{Yuying Sun}{Boston University}
\icmlauthor{Tianmin Shu}{Johns Hopkins University}
\icmlauthor{David Abel}{Google DeepMind}
\end{icmlauthorlist}

\icmlaffiliation{MIT}{Massachusetts Institute of Technology}
\icmlaffiliation{Boston University}{Boston University}
\icmlaffiliation{Johns Hopkins University}{Johns Hopkins University}
\icmlaffiliation{Google DeepMind}{Google DeepMind}

\icmlcorrespondingauthor{Andi Peng}{andipeng@mit.edu}

\icmlkeywords{Machine Learning, ICML}

\vskip 0.3in
]

\printAffiliationsAndNotice{}  % leave blank if no need to mention equal contribution

\begin{abstract}
Humans use social context to specify preferences over behaviors, i.e. their reward functions.
Yet, algorithms for inferring reward models from preference data do not take this social learning view into account.
Inspired by pragmatic human communication, we study how to extract fine-grained data regarding \textit{why} an example is preferred that is useful for learning more accurate reward models.
We propose to enrich binary preference queries to ask both (1) which features of a given example are preferable in addition to (2) comparisons between examples themselves.
We derive an approach for learning from these feature-level preferences, both for cases where users specify which features are reward-relevant, and when users do not.
We evaluate our approach on linear bandit settings in both vision- and language-based domains. Results support the efficiency of our approach in quickly converging to accurate rewards with fewer comparisons vs. example-only labels. 
Finally, we validate the real-world applicability with a behavioral experiment on a mushroom foraging task. Our findings suggest that incorporating pragmatic feature preferences is a promising approach for more efficient user-aligned reward learning.
\end{abstract}

\input{1-introduction}
\input{2-related_work}
\input{3-theory}
\input{4-approach}
\input{5-experiments}
\input{6-user_study}
\input{7-discussion}
\input{10-impact_statement}
\input{9-acknowledgements}

\clearpage
\bibliography{bibliography}
\bibliographystyle{icml2024}

\newpage
\input{8-appendix}

\end{document}

%% file: packages.tex
\usepackage{microtype}
\usepackage{graphicx}
\usepackage{subfigure}
\usepackage{booktabs}
\usepackage{hyperref}

\usepackage{algorithm, algpseudocode}
\usepackage{setspace}
\usepackage{amsmath}
\usepackage{amssymb}
\usepackage{mathtools}
\usepackage{amsthm}
\usepackage[shortlabels]{enumitem}
\usepackage[dvipsnames]{xcolor}

\usepackage[capitalize,noabbrev,nameinlink]{cleveref}
\creflabelformat{equation}{#2#1#3}
\usepackage{csquotes}
\usepackage{multirow}
\usepackage{mathrsfs}
\usepackage{mdframed}
\usepackage{bm}
\usepackage{float}
\usepackage{bbm,tipa}
\usepackage{mathbbol}
\DeclareSymbolFontAlphabet{\mathbb}{AMSb}
\DeclareSymbolFontAlphabet{\mathbbl}{bbold}
\usepackage{inconsolata}

\usepackage[accepted]{icml2024}

\usepackage[capitalize,noabbrev]{cleveref}

\theoremstyle{plain}

\theoremstyle{definition}

\theoremstyle{remark}

\usepackage[textsize=tiny]{todonotes}

%% file: commands.tex
\newcommand{\mc}[1]{\mathcal{#1}}

\newcommand{\mb}[1]{\mathbbm{#1}}
\newcommand{\mr}[1]{\mathrm{#1}}

\newcommand{\ra}{\rightarrow}

\def\reaches{\rightsquigarrow}

\newcommand\alwaysreach{%
  \mathrel{\ooalign{\hss$\square$\hss\kern-0.22ex\hbox{{$\reaches$}}}}}
\usepackage{calligra}
\DeclareMathAlphabet{\mathcalligra}{T1}{calligra}{m}{n}
\usepackage{etoolbox}
\makeatletter
\patchcmd{\NAT@test}{\else \NAT@nm}{\else \NAT@hyper@{\NAT@nm}}{}{}
\makeatother

\newcommand{\rlhf}{\textcolor{Black}{\textbf{RLHF}}}
\newcommand{\fp}{\textcolor{ForestGreen}{\textbf{FP}}}
\newcommand{\rlhfprag}{\textcolor{Cyan}{\textbf{Prag RLHF}}}
\newcommand{\fpprag}{\textcolor{RubineRed}{\textbf{Prag FP}}}

%% file: 1-introduction.tex
\section{Introduction}
\label{sec:intro}

\begin{figure*}
    \centering
    \includegraphics[width=0.9\textwidth]{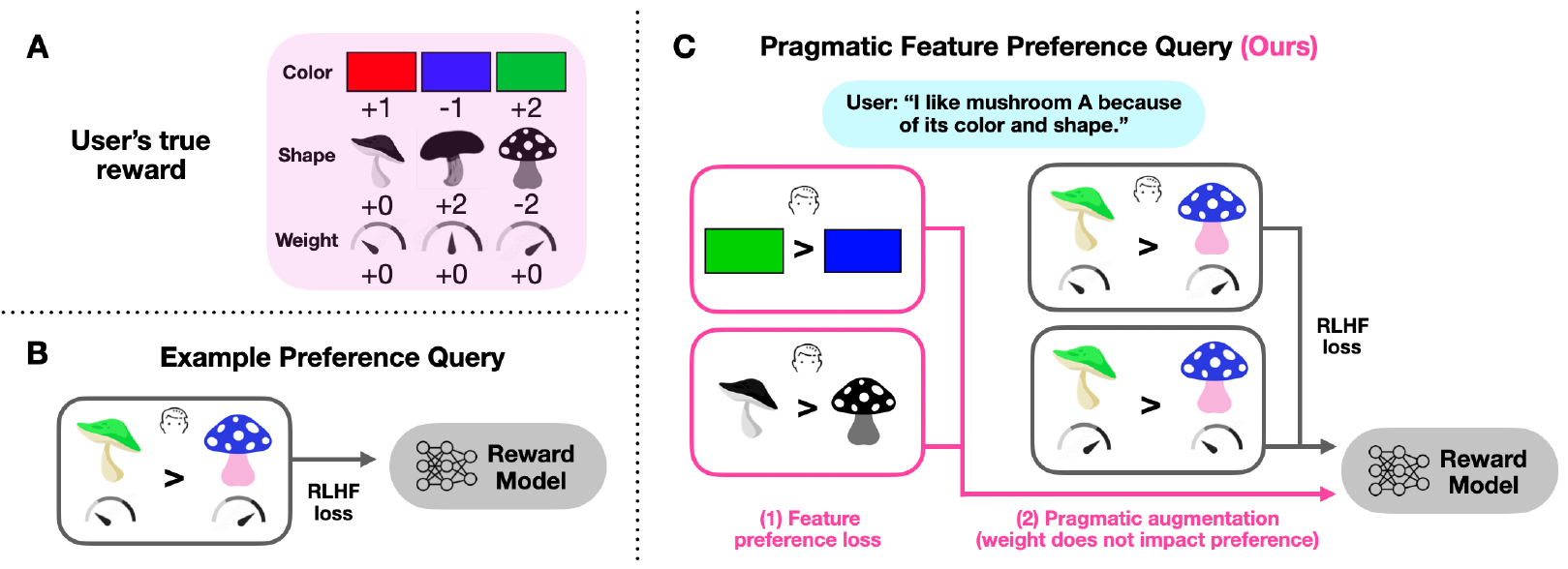}
    \caption{\textbf{A.} An illustrative user reward function in the mushroom foraging task. Rewards are a linear combination of color, shape, and weight features. \textbf{B.} \textit{Example preference} queries learn a traditional RLHF loss over example-level comparisons. \textbf{C.} Our approach, \textit{pragmatic feature preference} queries, makes use of (1) fine-grained feature-level preferences in conjunction with example-level preferences, and (2) language descriptions to infer reward-relevant features and augment preference data.}
    \label{fig:front}
\end{figure*}

Learning user-aligned reward functions from human data is a cornerstone of efforts in value alignment and AI safety \citep{fisac2020pragmatic,amodei2016concrete,christian2021alignment,hadfield2017inverse}.
Current efforts such as reinforcement learning (RL) from human feedback (RLHF) propose to learn reward functions from pairwise comparisons provided by human users \citep{christiano2017deep,griffith2013policy}.
Motivated by the idea that pairwise comparisons are a relatively simple and easy way for users to provide offline input for training a reward model, RLHF approaches have been used to train more efficient robotic systems \citep{basu2018learning,hullermeier2008label,jain2015learning}, and safer language models (LMs) \citep{bai2022training,bai2022constitutional}.
Unfortunately, because such feedback is provided over example pairs, valuable information regarding fine-grained components of the reward, i.e.\ \textit{which} features of the examples matter and \textit{why}, are lost \cite{basu2018learning}.

As a simple example, consider taking up the task of mushroom foraging introduced by \citet{sumers2022talk} (Fig.~\ref{fig:front}). How might we learn which mushrooms are good for foraging?
A pairwise comparison between two examples may tell us that one mushroom is better (that is, more delicious) than the other, but not the reason why (green mushrooms tend to be zestier in flavor).
Moreover, users may not hold the same preferences over which \textit{features} of mushrooms are important---a chef may prefer mushrooms to taste delicious but a collector may instead prefer them to look exotic.
In other words, there may be different \textit{reward-relevant features} that shape each user's \textit{preference relation} such that their underlying reward functions are different.

If we assume the user in question is not simply acting as an oracle providing labels divorced from the learning process, but rather as an engaged cooperative agent capable of providing descriptive feedback, we can treat users as active \textit{teachers} capable of providing richer information regarding their underlying reward function.
Such pedagogical models have been found to be useful for guiding RL agents from actions \citep{ho2016showing,goyal2018} and language feedback \citep{bisk2016natural,sumers2022talk,lin2022inferring}.
How might we do the same for preference learning?

In this work, we propose a pedagogical framework for modeling feature-level pairwise comparisons and design a joint loss to learn rewards from both feature and example-level comparisons.
Our key insight is that humans communicate preferences \textit{pragmatically}: when they describe which features of each example are important to their preference, they are also implicitly revealing which features \textit{are not important}.
For example, as shown in \cref{fig:front}, the fact that a user prefers a mushroom because of its color and shape might implicitly reveal that they do not care about mushroom weight or that weight does not matter for their preference.
This information can be used to expand the existing comparison-level data greatly, e.g., the user should hold the same preference over these two mushrooms \textit{even if their weights were flipped}. We introduce this pedagogical approach as learning from \textit{pragmatic feature preferences}.

First, we formalize the relationship between preferences over examples and preferences over features in a linear bandit setting. We propose a method to query for feature-level as well as example-level preferences and define a joint loss for learning from such input.
Second, we contribute a pragmatic approach for making additional use of this data by performing feature-level augmentation of \textit{non-relevant} reward features from linguistic preference descriptions. 

We evaluate our approach in experiments in both the mushroom foraging task (a vision-based domain) and a flight booking task (a language-based domain) \citep{lin2022inferring}.
We find that learning from pragmatic feature preferences outperforms baselines that only learn from either only example-level preferences or only pragmatic-augmented features, verifying that both elements are important for making use of contextual information contained in preference descriptions.
Importantly, we verify in a user study that such rich queries \textit{do not} significantly increase user effort with providing labels compared to RLHF.
Overall, our findings suggest that incorporating models of pragmatic human communication is important for efficient user-aligned reward learning.

\footnotetext{Code available at \\ \href{https://github.com/andipeng/feature-preference}{\texttt{github.com/andipeng/feature-preference}}}

%% file: 2-related_work.tex
\section{Related Work}
\label{sec:lit}
Traditional RL assumes that the reward function is given to a decision-making agent \citep{sutton1992introduction}, a practice that is subject to value misalignment and misspecification \citep{amodei2016concrete}. Ergo, a growing body of work proposes to instead infer the reward function from human data.

\textbf{Learning from demonstrations (IRL)}. 
Inverse reinforcement learning (IRL) methods propose to learn the reward function from observed actions in the environment, e.g. human demonstrations \citep{ng2000algorithms,Abbeel2004,ziebart2008maximum}. 
Unfortunately, such methods suffer from identifiability issues \citep{ziebart2008maximum,sumers2022talk}. That is, multiple reward functions can explain the same observed behavior. 
Moreover, IRL suffers from strong assumptions regarding the optimality of the demonstrator, or in other words, that the observed actions are always optimal under the user's true reward.

\textbf{Learning from pairwise preferences (RLHF).} 
With the rise of language models (LMs), there is renewed interest in learning rewards from pairwise preferences, colloquially referred to as reinforcement learning from human feedback (RLHF) \citep{christiano2017deep,griffith2013policy}. 
Motivated by the idea that binary preference labels are less burdensome for human users to provide, RLHF has emerged as a popular method for fine-tuning LMs \citep{kaufmann2023survey,wu2023fine}, although there are open questions regarding its efficiency and accuracy of reward modeling to true human preferences \citep{casper2023open}.

\textbf{Learning from teachers (pedagogy).}
Unlike the above approaches which assume data is generated by a user that is merely \textit{showing} what the correct thing to do is, pragmatic approaches instead incorporate models of users that are \textit{teaching} \citep{ho2016showing,sumers2022talk,lin2022inferring} why this is the correct thing to do.
This subtle distinction manifests in algorithms that explicitly incorporate pedagogical models, i.e. models where human-generated data is intentionally intended to be informative about the user's underlying reward function \citep{hadfield2017inverse,fisac2020pragmatic}.

\textbf{Learning with state abstraction.} There is substantial evidence to suggest much of the generalizability of human learning and planning can be attributed to abstraction, i.e.\ the selective filtering of task-relevant information \citep{ho_val_abstr2019,ho2022people}.
This suggests that flexibly creating abstractions containing task-relevant features is important to downstream generalizable learning, particularly with limited examples \citep{peng2024learning,bobu2023aligning}. 

In this paper, we unify different streams of work in pedagogical reward learning and human abstraction to develop a model of learning from pairwise preferences that takes into account human input that is explicitly informative of task-relevant features.
Such a framework offers two benefits: first, targeting preference data to learn rewards at the feature-level enables more efficient learning given limited comparisons; second, humans can provide descriptive feedback on important features in language, offering a more natural teaching process.
In the next section, we explore how both can be utilized to learn better reward models.

%% file: 3-theory.tex
\section{Preliminaries}
\label{sec:theory}

Our primary focus is on the \textit{reward modeling problem} in which we seek to learn a reward function that aligns with a user's unknown preference relation while observing only finitely many comparisons from that preference relation.

% Contextual Bandits and reward modeling.
%
We study reward modeling in contextual bandit problems \cite{langford2007epoch,lattimore2020bandit}, which are a middle point between $k$-armed bandits and full sequential decision-making problems. A contextual bandit presents a challenging decision-making problem due to both the explore-exploit dilemma and generalization but does not introduce the complexities of credit assignment and long-term planning. For this reason, it is a compelling choice for studying preference-based reward modeling.

% Contextual bandits.
%
\vspace{-1em}
\paragraph{Contextual Bandits.} A contextual bandit in its general form is a model of a decision-making problem defined by the tuple $(\mc{C}, \mc{A}, \mu, R)$, where $\mc{C}$ is a set of contexts, $\mc{A}$ is a set of actions, $\mu \in \Delta(\mc{C})$ is a probability distribution over $\mc{C}$, and $R : \mc{C} \times \mc{A} \ra \mathbb{R}$ is a reward function. We note that while the reward function is typically stochastic in most bandit problems, in the setting we study, the reward function is deterministic. At each time step, a context $c \in \mc{C}$ is sampled $c \sim \mu$ and presented to the decision-maker. The decision-maker then chooses an action $a \in \mc{A}$ and observes $R(c,a)$ with the goal of maximizing some measure of long term reward. 
We follow the conventions of \citet{sumers2022talk} and study a special case of linear contextual bandits \cite{li2010contextual} in which each context is a subset of the action space that the agent is allowed to choose from in that context. For instance, in the mushroom foraging task, each context is a collection of mushrooms the agent must choose from. More formally, the action space is the set of $n$-dimensional vectors, $\mc{A} = \mathbb{R}^n$, and each context is simply a subset of this space, $c \subseteq \mc{A}$. The agent is then only allowed to choose an action contained in the current context, and the reward function is only well-defined for cases where $a \in c$. In such cases, it is sufficient to express the reward function as only a function of $a$, $R: \mc{A} \ra \mathbb{R}$.

% Reward Modeling.
%
\vspace{-1em}
\paragraph{Reward Modeling.} In a contextual bandit of the kind described above, the reward modeling problem is defined as follows. We are given as input the context set $\mc{C}$, the action space $\mc{A}$, and a finite set of preference data over actions $\mc{D} = \{(a_i, a_i', f(a_i, a_i'))\}_{i=1}^m$, where $a_i, a_i' \in \mc{A}$ are each actions, and $f : \mc{A} \times \mc{A} \ra \{\succ, \prec, \sim\}$ is a function mapping each action pair to a preference relation. We suppose the preference relation is unknown, and wish to learn a reward function, $\hat{R} : \mc{A} \ra \mathbbm{R}$ that aligns with the underlying preference relation that generated the preference data. Notice that since each context is simply a subset of the action space, the preference relation of interest is over pairs of actions, and the reward function we wish to learn is also a function of action, rather than a context-action pair. 

Following previous work in IRL \cite{sumers2022talk}, we assume that the reward $\hat{R}$ (e.g. tastiness of a mushroom) is a linear combination of feature rewards $\hat{R}^j$ (e.g. tastiness of a green mushroom), so that: $\hat{R}_\theta(a) = \sum_{j=1}^n \theta^j \hat{R}^j(a^j)$, where $a^j$ is the value associated with a specific feature (e.g. green), and $\theta^i$ is the $i$-th element of a linear weight vector on feature rewards. When clear from context, we abbreviate $\hat{R}^j(a^j)$ to simply $\hat{R}(a^j)$. 
The traditional goal is then to learn a $\theta$ such that $\hat{R}_\theta(a_1) > \hat{R}_\theta(a_2)$ if and only if $f(a_1, a_2) =\ \succ$.

% Feature preferences.
%
We propose to consider pairwise \textit{feature preferences} over different settings of an individual feature of each action. For instance, consider two actions comprised of three features, $a_1 = \langle 0, -1, 2 \rangle$ and $a_2 = \langle 1, 0, 0 \rangle$.  We refer to the $j$-th feature of action $a_2$ as $a_2^j$. We then let $\phi : \mb{R} \times \mb{R} \ra \{\succ, \prec\}$ express a \textit{feature preference} relation, indicating whether the value of the $j$-th feature of one action is preferred to another.

For example, consider two actions $a_1 = \langle 1, 0, 20 \rangle$, and $a_2 = \langle 5, 2, 12 \rangle$ each describing a mushroom. Suppose the first feature ($a_1^1 = 1, a_2^1 = 5$) captures the zestiness of the mushroom. A user that dislikes zesty might be thought of as maintaining the feature preference relation $\phi(a_1^1, a_1^2) =\ \succ$. As a shorthand, we denote such outcomes as $a_1^1 \succ_\phi a_1^2$.

% Relating feature prefs w context prefs.
%
Our assumption is that an individual's preference about the \textit{features} of an object will inform their overall preferences regarding that object.
%
% Summary?
%
Our primary hypothesis is that decomposing a preference relation about a pair of objects into preferences about the \textit{features} of those objects allows for more effective reward modeling. It is worth noting that there are situations where the assumptions introduced thus far don’t hold, such as when it is impossible to decompose an example-based reward into its constituent feature-based rewards. Such situations may arise in scenarios where humans do not hold preferences over features of an object independent of the object itself (for example, a human may prefer a football to a basketball, and otherwise not care about individual features of balls such as bounciness, color, size, etc.). In our experiments, we study some deviations from these assumptions and acknowledge that a full analysis of how our method accommodates these more general settings is an important direction for future work.

%% file: 4-approach.tex
\section{Approach: Pragmatic Feature Preferences}
\label{sec:approach}

% Framing.
%
To address the reward modeling problem, our primary assumption is that any individual's preference relation about elements of a given domain is tightly coupled with \textit{how} that individual represents elements from that domain. For example, suppose an individual were to prefer a zesty mushroom to a mild mushroom---if zestiness is a primary determining factor in a person's preference about mushrooms, it is likely that zestiness is directly represented by that person, too. This assumption unlocks two key elements.

% First: Feature-level comparisons.
%
\vspace{-1em}
\paragraph{Element 1: Feature-level comparisons.} First, we can solicit extra preference information from users as \textit{feature-level comparisons}, rather than solely at the example level. In the mushroom case, this means we can simply ask whether someone prefers spicy to non-spicy foods, rather than ask which of the two mushrooms they prefer. We formalize this below by forming a joint loss term that balances between feature-level comparisons and example-level comparisons.

% Second: Pragmatic data augmentation.
%
\vspace{-1em}
\paragraph{Element 2: Pragmatic data augmentation.} Second, we can infer which features are \textit{unimportant} to the user's preference in order to significantly expand the available labeled preference data. For instance, if we ask a user to point out which features are most significant for deciding between two mushrooms and they respond with ``spice level" and ``color", we suggest it is natural to infer that the other mushroom features are \textit{unimportant} for the given comparison, and consequently we can synthesize new training data where the unimportant feature values are swapped while preserving the object-level preference relation. We provide more concrete details below.

% --- Feature-level queries: Augmented Loss ---
\subsection{Feature-level queries: Enriched Loss}

First, we enrich the preference data collected by not only capturing example-level comparisons but also feature-level comparisons. For example, in the mushroom domain introduced by \citet{sumers2022talk}, each mushroom is associated with some features such as its size and color. In such a case, we can ask users: (1) Do you prefer mushroom A or mushroom B?, and (2) Do you prefer the size of mushroom A or mushroom B? Do you prefer the color of mushroom A or mushroom B? These fine-grained queries are intended to extract additional information per example pair that can be used to train a reward model.

% RLHF Loss.
%
\paragraph{RLHF Loss.} More formally, we adopt the standard conventions of RLHF in which the learned reward model, $\hat{R}$, is chosen to minimize the cross-entropy between the reward model's predicted preference labels and the actual labels provided by the user, following the Bradley-Terry model which states humans are noisily rational in identifying the correct example \citep{bradley1976science} :
\begin{align}
    \label{eq:rlhf-loss}
    \texttt{rlhf}&\texttt{-loss}(\hat{R}, \mc{D}) = - \sum_{(a_1, a_2, f) \in \mc{D}} \\
    &\left(\mathbbm{1}_{a_1,a_2}^f \log \hat{P}(a_1 \succ a_2)\ +
    \mathbbm{1}_{a_2, a_1}^f \log \hat{P}(a_1 \prec a_2)\right). \nonumber
\end{align}
where $\mathbbm{1}_{a_i,a_j}^f$ expresses the indicator function on whether $a_i \succ_f a_j$, and $\hat{P}(a_1 \succ a_2)$ is the learned reward function's inferred preference over $(a_1, a_2)$ as defined by the ratio:
\begin{equation}
\label{eq:exponentiated_reward}
    \hat{P}(a_1 \succ a_2) = \frac{\textrm{exp}(\hat{R}(a_1))}{\textrm{exp}(\hat{R}(a_1))+\textrm{exp}(\hat{R}(a_2))}.
\end{equation}

% Feature-loss.
%
\paragraph{Feature-Pairwise Loss.} We propose to enrich this loss with a feature-level loss following the same convention. That is, given two actions, $a_1$ and $a_2$, where the features of the first action are all preferred to the second,
\begin{align}
    % feat loss text.
    %
    \texttt{feat}&\texttt{-loss}(\hat{R}, \mc{D}) = - \sum_{(a_1,a_2,\phi \in \mc{D})} \sum_{j=1}^n \\
    %
    % Indicator
    &\left(\mathbbm{1}_{a_1^j, a_2^j}^\phi \log \hat{P}(a_1^j \succ a_2^j) + \mathbbm{1}_{a_2^j, a_1^j}^\phi \log \hat{P}(a_1^j \prec a_2^j)\right).\nonumber
\end{align}
Again, $\mathbbm{1}_{a_1^j, a_2^k}^\phi$ denotes the indicator function on whether $a_1^j \succ_\phi a_2^k$, and $\hat{P}(a_1^j \succ a_2^j)$ is the ratio for the learned reward function's output of feature $j$:
\begin{equation}
    \hat{P}(a_1^j \succ a_2^j) = 
    \frac{\exp(\hat{R}(a_1^j))}{\exp(\hat{R}(a_1^j)) + \exp(\hat{R}(a_2^j))}.
\end{equation}
Again, we note that this is where we exploit the linearity assumption---we assume that all reward models of interest $\hat{R}$ can compute a per-feature reward, $\hat{R}(a_i^j$), for any action $a_i$ and feature $j$.

Our overall loss is then simply a weighted sum of the two,
\begin{align}
    \texttt{loss}&(\hat{R}, \mc{D}) = \\
    %
    % RHS.
    &(1-\beta) \texttt{rlhf-loss}
    (\hat{R}, \mc{D}) +
    \beta \texttt{feat-loss}(\hat{R}, \mc{D}), \nonumber
\end{align}
with $\beta \in [0,1]$ a hyperparameter that trades off between the strength of the feature pairwise loss (\texttt{feat-loss}) and the example pairwise loss (\texttt{rlhf-loss}).% 

% --- Pragmatic Data Augmentation ---
\subsection{Pragmatic Data Augmentation}

The second consequence of asking a user for their feature preferences is that we can also ask them to describe features that are \textit{important} for determining their overall preference. By doing so, we can isolate which features contribute to their preference between the two examples, and thus also infer which features are irrelevant for determining their preference.\footnote{We refrain from also asking users about which features are irrelevant, both due to the redundancy of the query and the potential for a high number of irrelevant features. However, non-pragmatic feature preference queries can learn from a full set of feature preference labels if available.} One immediate benefit of knowing which features are irrelevant to a user in forming a specific preference is that we can expand the available labeled preference data by synthesizing new data points where the preference label remains the same, but the irrelevant features are modified.

Concretely, when we query a user for their choice between $a_1$ and $a_2$, we further ask a user to describe, in language, what features are important for making this decision. We then infer that any feature not mentioned is irrelevant for determining preference, and synthesize a new data point for each possible swap of the irrelevant features' values.

% Simple example.
%
For example, consider a simple case where each action is characterized by only two features, and we query a user on the pair $a_1 = \langle 0, 1 \rangle$ and $a_2 = \langle 10, 2 \rangle$. Suppose the user prefers $a_1$ to $a_2$, and indicates that the first feature was most important for determining their preference (perhaps this captures the potential poison content of a mushroom). Then, we synthesize a single new data point by keeping the preference label but swapping the irrelevant features. So, we construct $a_1^\circ = \langle 0, 2 \rangle$ and $a_2^\circ = \langle 10, 1 \rangle$, and assert that $a_1^\circ \succ_f a_2^\circ$. Naturally, with only two features the amount of synthesized data is minimal, but as the number of total features increases, the opportunity for this approach to improve training speed increases as well.

% Algorithm.
%
The pseudocode for carrying out this pragmatics-inspired preference augmentation is given in \cref{alg:augment_data}. The $\texttt{mask}$ function is assumed to set the inferred irrelevant features to a special character, ``$\emptyset$", and $\texttt{feat-combos}$ constructs the set of all combinations of indices of the features set to $\emptyset$. We then use the notation $\vec{j}$ to refer to a vector of indices, and $a_1^\circ[\ \vec{j}\ ] = a_2[\ \vec{j}\ ]$ as shorthand to refer to assigning each feature with an index contained in $\vec{j}$ to its value in $a_2$. We provide an expanded pseudocode that adds additional clarity in the Appendix as Algorithm 2. 
%
% Two kinds of augmentation.
It is worth noting that there is an important subtlety to how we approach this data augmentation that depends on how we implement \texttt{feat-combos}. In the first method of implementation, we construct all possible combinations of new data where the irrelevant features take on \textit{any possible value}. In the second, we only construct combinations that can result from swapping the feature values that are seen in the specific data point. These two methods each make different assumptions about the underlying pragmatic inference: the first method assumes that \textit{the features inferred to be irrelevant are irrelevant in general}, whereas the second method assumes that \textit{the features are inferred to be irrelevant in this specific context, for these specific values}. We make this second assumption as it is a more cautious approach to data augmentation, but note that exploring the general difference between the two methods could be a useful direction.

\input{algorithms/data_augment}

%% file: algorithms/data_augment.tex
\algtext*{EndWhile}% Remove "end while" text
\algtext*{EndIf}% Remove "end if" text
\algtext*{EndFor}% Remove "end if" text
% ------------------
% -- Augment Data --
% ------------------
\begin{algorithm}[!t]

\caption{Pragmatic Feature Preference Augmentation}
\label{alg:augment_data}
\vspace{1mm}
\textsc{Input:} $\mc{D}$ the preference dataset. \\
\textsc{Output:} $\mc{D}'$ the augmented preference dataset. \\

\begin{algorithmic}[1]
%\setstretch{1.2}
%
\State \textbf{Init:} $\mathcal{D'}=\mathcal{D}$
\For{$(a_1, a_2, f(a_1,a_2), \texttt{mask}) \in \mc{D}$}
\State $a_1^\circ, a_2^\circ = \texttt{mask}(a_1, a_2, f(a_1, a_2))$
\For{$\vec{j} \in \texttt{feat-combos}(a_1^\circ, a_2^\circ)$}
\State $a_1^\circ[\ \vec{j}\ ] = a_2[\ \vec{j}\ ]$
\State $a_2^\circ[\ \vec{j}\ ] = a_1[\ \vec{j}\ ]$
\State $f^\circ = f(a_1, a_2)$
\State $\mc{D}' = \mc{D}' \cup \{(a_1^\circ, a_2^\circ, f^\circ)\}$
\EndFor
\EndFor
\Return $\mc{D}'$
\end{algorithmic}
\end{algorithm}
\setlength{\textfloatsep}{15pt}
%%%%%%%%%%%%%%%%%%%%%%%%

%% file: 5-experiments.tex
\vspace{-1mm}
\section{Experiments}
\label{sec:experiments}

\begin{figure*}[t!]
    \centering
    \includegraphics[width=0.9\textwidth]{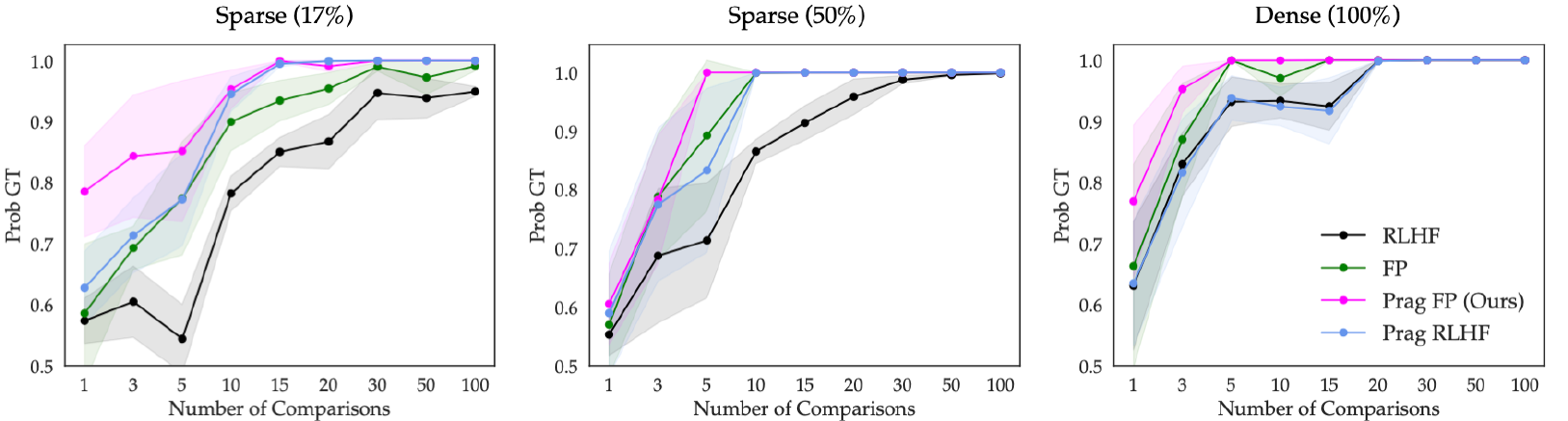}
    \vspace{-1mm}
    \caption{Results with simulated preference labels on the mushroom foraging task. \fpprag \ outperforms other methods, converging to a more accurate learned reward given fewer seen examples. This effect is especially prominent as the reward-relevant features become more sparse. Confidence bounds depict standard error across five independent seeds.}
    \vspace{-1em}
    \label{fig:sim_mushrooms}
\end{figure*}

To validate our approach empirically, we conduct experiments in two domains: a vision-based mushroom foraging task and a language-based flight booking task.
We begin with experiments that simulate user preference responses based on some known ground truth reward functions to study the learning efficiency of feature preferences in the mushroom task, which is a domain that allows for direct control over the reward functions and their feature densities.
Second, we conduct experiments with real language descriptions collected in the flight booking task to explore the benefits of our pragmatic framework with linguistic data.

\textbf{Evaluation.}
Our goal is to learn accurate reward models that assign high rewards to actions that better satisfy a user's true reward function.
To measure the success of learned models, we evaluate the probability of the ground truth best examples \citep{basu2018learning}. %
The higher the probability assigned to the ground truth (i.e. the example that maximizes the true reward), the more accurate the learned reward parameters are.
We report results on five independently trained seeds.

\textbf{Implementation details.}
We implement all reward models as linear networks (single layer, no activations).
Each feature predictor in the joint model is trained independently without sharing parameters, and their resulting outputs are concatenated and fed through a final layer for reward prediction.
We swept possible $\beta$ values and found $0.5$ consistently achieved the best performance.

\subsection{Understanding reward sparsity's impact on learning efficiency}
We begin by testing the hypothesis that given perfect user labels, i.e.\ an oracle user that answers both example and feature preference queries along with providing reward-relevant features according to the ground truth, feature preference queries will learn more accurate rewards from less examples compared to baselines. 
In particular, we study how the two distinct elements of our approach---feature preferences and our pragmatic augmentation framework---are impacted by the sparsity of reward features. That is, we explore how the percentage of task-relevant reward features that characterize the ground truth preference relation will impact the quality of the learned reward given a fixed budget of example pairs.

\textbf{Task 1: mushroom foraging.}
To disentangle these two factors, we make use of a highly controlled task where we can design different types of ground truth preference relations in terms of the types of reward functions used to represent these preferences. Inspired by \citet{sumers2022talk}, we create a vision-based task where users play the role of a mushroom forager in charge of teaching which mushrooms are preferred. 
Mushrooms are parameterized by six possible discrete features: \textit{texture, color, shape, height, weight} and \textit{smell} with three possible values for each feature (e.g. \textit{stinky, pleasant,} and \textit{neutral} for smell).
We generate reward functions of three different kinds, each characterized by a parameter vector $\theta_{\mr{GT}} \in \{-2, -1, 0, 1, 2\}^6$: (1) \textbf{dense (100\%)} (all six features are reward-relevant), (2) \textbf{sparse (17\%)} (only a single feature is reward-relevant), and (3) \textbf{sparse (50\%)} (three features are reward-relevant). 
For each reward type, we generate two reward functions by randomly sampling the subset of features that are task-relevant, and then randomly sampling the value of each feature.

%\textbf{Comparisons.}
User queries consist of a task, a reward function, and a randomly sampled comparison (see \cref{fig:front}.)
For each query, we change the type of labels collected for learning: (1) comparison (\rlhf, baseline) queries use example-level comparisons only, (2) feature preference (\fp, ablation) queries use feature-level in addition to example-level comparisons, (3) pragmatic comparison queries (\rlhfprag, ablation) use linguistic utterances describing reward-relevant features in addition to example-level comparisons, and (4) pragmatic feature preference (\fpprag, our approach) queries combine the pragmatics augmentation framework in conjunction with feature- and example-level preference comparisons.
Queries return 1 if A is preferred to B, and $-1$ otherwise.

\textbf{Results.} As shown in \cref{fig:sim_mushrooms}, our results indicate that \fpprag \ converge to a more accurate learned reward with fewer examples required compared to other approaches.
Across all three types of reward functions, we find that \fp \ as well as \rlhfprag \ contribute meaningfully to learning efficiency, particularly in low-example regimes.
When we remove either method, we see performance slightly falter when compared to combining both, highlighting their combined value.
\rlhf \ performs the worst, indicating that valuable information is lost by modeling the problem solely over example-level comparisons without context.

%\textbf{Reward Sparsity.}
We further evaluate the quality of the learned reward model based on the sparsity of the reward function generated. This is motivated by the belief that real-world rewards are generally feature sparse \citep{bajcsy2018learning}---that is, users hold preferences based on a few, not many, task-relevant features, 
As seen across each plot in \cref{fig:sim_mushrooms}, the magnitude of improvement is especially apparent in the more sparse reward features, confirming the hypothesis that pragmatics-motivated fine-grained feedback is most advantageous when few features impact the final preference relation.% We suggest this is a useful finding to motivate further research into larger scale domains.

\subsection{Analyzing the impact of linguistic descriptions}
In the previous experiment, we simulated perfect knowledge of reward-relevant features.
Now that we have established the value of including both the pragmatics framework in conjunction with feature-level preferences, we next explore how real linguistic descriptions impact learning.

\textbf{Task 2: flight booking.}
To study more natural linguistic input, we require a domain where large amounts of real human linguistic descriptions are collected.
With this in mind, we use a language-based task from \citet{lin2022inferring} where users play the role of a flight booker in charge of teaching which flights are preferred. 
Flights are parameterized by eight possible features: \textit{arrival time before meeting, american, delta, jetblue, southwest, longest stop, number of stops} and \textit{price} (airline features are discrete whereas the rest are continuous). 
Reward functions are randomly generated, where $\theta_{GT} \in \{-1, -0.5, 0, 0.5, 1\}^8$. Importantly, \citet{lin2022inferring} collected a human dataset where these reward functions are paired with human linguistic utterances generated by real users describing their preferences over those rewards in language.\footnote{The original dataset can be found at \href{https://github.com/jlin816/rewards-from-language}{\texttt{github.com/jlin816/rewards-from-language}}. Note, \citet{lin2022inferring} presented a set of three options to users and asked for a description of the option that was most optimal under the reward function, which we converted into two pairwise comparisons. The linguistic data was also collected over iterative feedback rounds to study the effect of recursive reasoning, which we disregard.}
Some example descriptions include ``american or delta prefered. more stops good, but long length of stops bad'' and ``i want the longest stop and the fewest number of stops''. We randomly sample 20 reward functions and their corresponding descriptions from the full dataset.

%\textbf{Turning utterances into feature maps.}
To make use of linguistic descriptions, we must convert unstructured linguistic utterances into structured feature maps specifying reward-relevant features.
In the simplest case, we could require the user to specify the reward-relevant features from the full list, but doing so requires additional human data collection and is subject to misspecification \citep{peng2023diagnosis}.
Therefore, we deployed a language model (LM) to parse descriptions into structured feature representations. Specifically, we prompt GPT-4 \cite{achiam2023gpt} with the linguistic description and full feature space to generate a feature map $\in \{0,1\}^8$ specifying reward-relevant features.

\begin{figure}
    \centering
    \includegraphics[width=.3\textwidth]{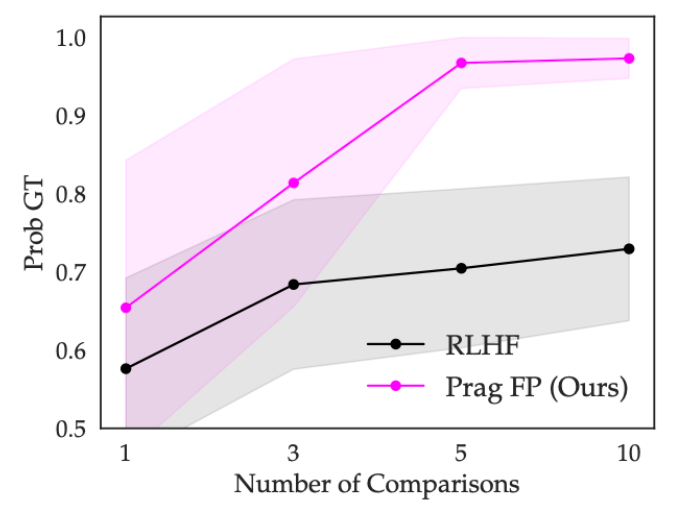}
    \caption{Results with real user descriptions on the flight booking task across 20 randomly sampled reward functions. Confidence bounds depict standard error across five independent seeds.}
    \label{fig:flights}
\end{figure}

\begin{figure*}
    \centering
    \includegraphics[width=0.9\textwidth]{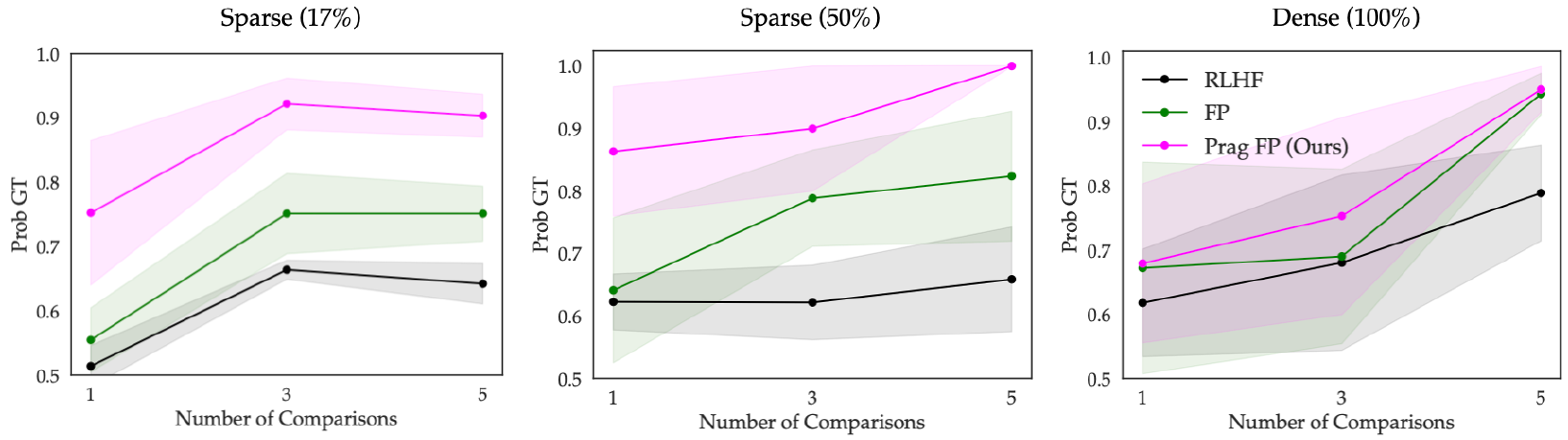}
    \vspace{-1em}
    \caption{Results with real user responses on the mushroom foraging task. These results corroborate the simulated results from \cref{fig:sim_mushrooms}. Confidence bounds depict standard error across five independent seeds.}
    \vspace{-1em}
    \label{fig:user_results}
\end{figure*}

\textbf{Results.}
As shown in \cref{fig:flights}, even with messy real linguistic data, \fpprag \  outperforms traditional \rlhf, converging to a more accurate reward with fewer examples required. We report results evaluated over 20 randomly sampled reward functions, each with five independent seeds. 

We note that we did not attempt to perform high robustness prompt engineering \citep{chen2023unleashing}, nor explicitly study question-answering mechanisms to elicit more accurate linguistic utterances from human users, although improvements across both axes would certainly further improve the accuracy of the pragmatic feature preference modeling.

%% file: 6-user_study.tex
\section{User Study}
\label{sec:user}

In \cref{sec:experiments}, we evaluated our framework with simulated human preference labels.
We now expand on these results with a behavioral study conducted with real users on the same mushroom foraging task. 
We are interested in addressing two questions. 
First, do real user labels for \fpprag \ queries validate our simulated results? 
Second, do users exert significantly more effort when giving these queries compared to \rlhf \ pairwise comparisons?

\textbf{Study Overview.}
We conducted a between-subjects user study where participants were asked to play the role of a mushroom forager tasked with selecting tasty mushrooms. 
Users were trained to read reward functions and calculate tastiness scores (rewards), and then given mushroom pairs and asked to give preferences about these pairs according to provided reward functions. 
We asked participants to answer three types of preference queries: \rlhf \ queries (preferences over mushroom examples), \fp \ queries (preferences over mushroom features in addition to examples), and \fpprag \ queries (language descriptions in addition to preferences over examples and features). 
Each user answered 30 queries for the same six reward functions as in \cref{sec:experiments} (five per reward). 
We used the responses from each query type to train a reward model.
We additionally asked participants brief survey questions (\cref{tab:survey_questions}) regarding their perceived effort, performance, and frustration after queries. 
Responses were assessed on a Likert scale \cite{likert1932technique}, with 1 being the lowest (``strongly disagree") and 7 the highest (``strongly agree"). 
We also recorded total time spent on the task.

\textbf{Participants.}
We recruited 36 participants, 12 for each query type, from Prolific, an online crowdsourcing site. 
Participants were screened according to the following characteristics: hold above a 95\% approval rating, speak English as a primary language, and reside primarily in the United Kingdom or United States. 
We paid participants at a rate of \$16 per hour and rejected responses that were low-effort (e.g.\ left responses blank, repeated the same answer for all questions, etc.). Our study passed institutional IRB review.

\subsection{Learning with real human responses.}
We begin by analyzing the impact of learning reward models from real human preferences, i.e.\ labels that may be noisily generated from users. 
We conduct the same training protocol as in \cref{sec:experiments} using randomly sampled user responses to train reward models for each reward function. 
For each reward, we report five independent seeds.

\textbf{Results.}
As shown in \cref{fig:user_results}, \fpprag \ outperforms baselines, especially as the reward is more sparse. 
These results corroborate the simulated results from above and provide meaningful evidence that real users are able to generate both linguistic descriptions and preference responses that can be used to accurately train reward models.

\begin{table}[H]
\centering
\setstretch{1.1}
\begin{tabular}{l}
    
  \textbf{Survey Questions}\\
  \hline
  Q1. Choosing the best mushroom was challenging.\\
  Q2. I could accurately communicate the best mushroom.\\
  Q3. Describing my preferences was relatively easy.\\
  Q4. I was frustrated with providing labels.\\
  \hline
\end{tabular}
\caption{Post-user study survey questions. User responses are assessed on a 1 to 7 Likert scale (with 7 being ``strongly agree'').}
\label{tab:survey_questions}
\end{table}

\subsection{Understanding impact on user effort.}
To ensure that \fpprag \ does not significantly negatively impact the user data collection process, we assessed the survey responses collected from participants at the conclusion of the study.
Questions are shown in \cref{tab:survey_questions}.
We analyzed Likert ratings using one-tailed independent $t$-tests, where both \fp \ and \fpprag \ queries are compared to \rlhf \ queries.

\textbf{Results.}
First, there was no significant difference in whether participants felt they could communicate preferences accurately (Q2) (no significant difference for either query type ($t(11)=-1.81, 0.29, p=0.08,0.77$)).
Participants who answered \fpprag \ queries \textit{did not} find it more challenging to describe their preferences (Q2, $t(11)=0.16, p=0.43$), while participants who answered \fp \ queries \textit{did} find it more challenging ($t(11)=2.31, p=0.02$)). 
This supports the hypothesis that allowing users to provide descriptions of important features pragmatically is more natural than providing feedback on all features.
Importantly, providing \fpprag \ queries \textit{did not} cause users to experience more frustration with providing labels (Q4, $t(11)=-0.87, p=0.19$) compared to \fp \ ($t(11)=-3.07, p=0.01$).
Lastly, users who provided \fpprag \ queries \textit{did not} take significantly more time on the task ($t(11)= -0.24, p=0.41$) compared to \fp \ queries, who did take longer ($t(11)=-1.86, p=0.03$).

%% file: 7-discussion.tex
\section{Discussion}
\label{sec:discussion}

%\textbf{Summary.}
We study a new form of user query, pragmatic feature preferences, for use in learning reward models from fine-grained human input.
Our method relies on two key elements: first, that human preferences at the feature level are valuable for learning accurate reward functions from fewer provided examples, and second, that what humans choose to describe in language tells us important information regarding which features are reward-relevant in their preference relation.

%\textbf{Limitations and Future Work.} 
Conceptually, our model builds on a rich history of work in pragmatic reasoning by explicitly modeling humans as teaching when giving feedback. 
While we studied our learning in an entirely offline setting, there are exciting directions for incorporating recursive reasoning in developing models that learn to ask the right questions for further clarifying inference of feature preferences in uncertain settings \citep{li2023eliciting}. 
Moreover, we made the assumption that given language descriptions, we can ground the identified features from those utterances to the correct features in the state representation, an assumption that is challenging in practice due to ambiguity in grounding ambiguous descriptions to an agent's perceptual state \citep{harnad1990symbol}. 
Lastly, the data augmentation aspect of our pragmatic framework relied on the ability to easily swap non-reward-relevant features in the comparison examples, which may be challenging with text-based models (e.g. swapping the \textit{toxicity} in outputs). 
Nonetheless, we are excited about the promise of incorporating pragmatics-inspired models of human abstract reasoning to learning more user-aligned reward models.

%% file: 10-impact_statement.tex
\section{Impact Statement}
\label{sec:impact}

This paper presents work on how to better learn individualized reward functions from human input. While this work is intended to help learn more accurate user objectives, we did not discuss possible misuse associated with malicious actors teaching models dangerous features. We will leave such discussion open to the broader socio-technical community.

%% file: 9-acknowledgements.tex
\section{Acknowledgements}
\label{sec:acknowledgements}

We thank Jacob Andreas and Alexis Ross for their formative insights and discussions, as well as reviewing earlier drafts of the paper. We are grateful to Ted Sumers and Jessy Lin for helpful advice on the mushroom foraging and flight tasks, respectively. We additionally thank Mark Rowland and the rest of the Google DeepMind team who reviewed an early draft of the paper. Andi Peng is supported by a NSF Graduate Research Fellowship and Open Philanthropy.

%% file: 8-appendix.tex
\appendix
\onecolumn
\label{sec:appendix}

\section{User Study}
\begin{figure}[H]
   \centering
   \includegraphics[width=0.9\textwidth]{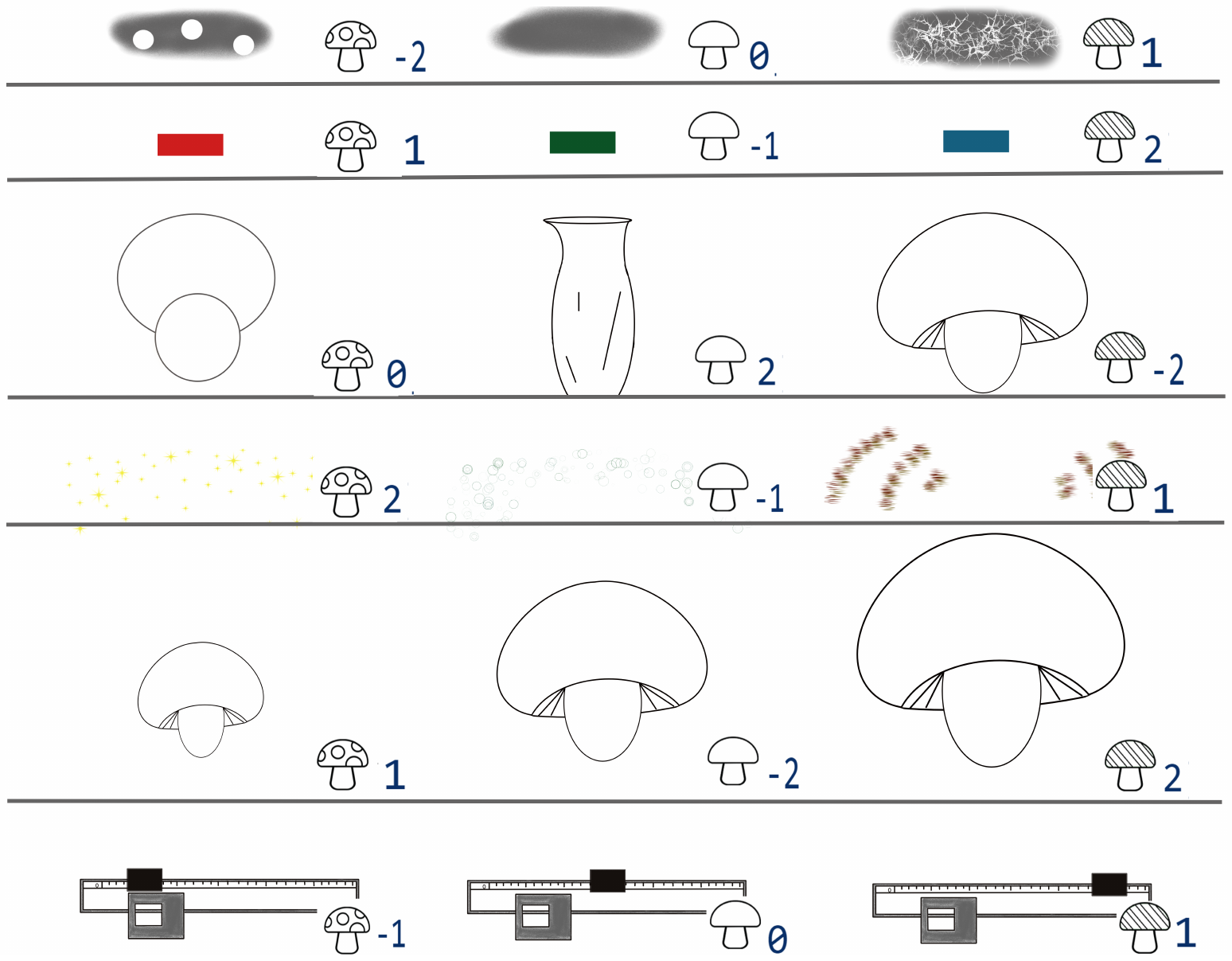}
   \caption{Example reward function provided in the user study.}
   \label{fig:example_map}
\end{figure}

\cref{fig:example_map} depicts one of the six reward functions presented in the user study. Users were trained to read reward functions in the familiarization stage of the study and then presented six unique reward functions to reference for queries.

\cref{fig:example_map} illustrate four possible mushrooms.

\begin{figure}[H]
   \centering
   \includegraphics[width=0.9\textwidth]{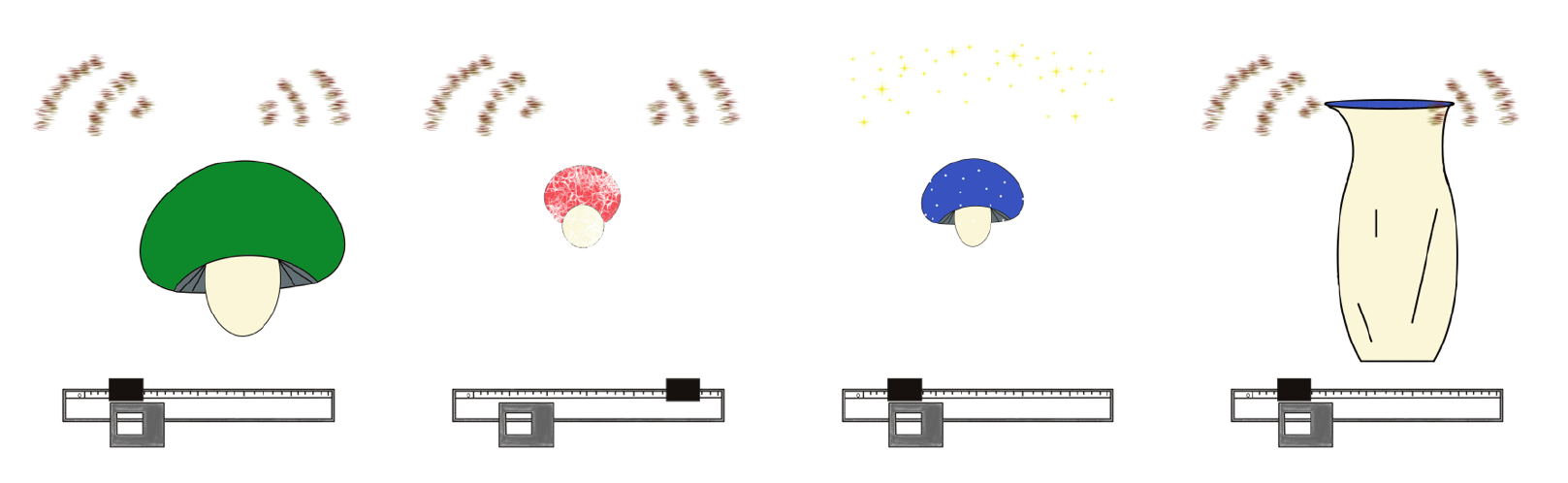}
   \caption{Four example mushrooms used in the user study.}
   \label{fig:example_mushrooms}
\end{figure}

\section{Expanded Pseudocode}

To remove any ambiguity about Algorithm 1, we here present a more detailed version of the pseudocode that makes the specifics of the algorithm more clear. We present this expanded version as Algorithm 2. Here, $\texttt{feat-combos}(a_1^\circ, a_2^\circ)$ constructs a set of ordered pairs of feature indices. This set, active-feature-pairs, contains pairs of features $(j_x, j_y)$ whose values should be swapped. The first element ($j_x$) is an index into $a_1$, and the second element ($j_y$) is an index into $a_2$. Lines 8 and 9 perform the swapping on copies of $a_1$ and $a_2$, while line 10 computes the new preference relation.

\input{algorithms/expanded_data_augment}

%% file: algorithms/expanded_data_augment.tex
\algtext*{EndWhile}% Remove "end while" text
\algtext*{EndIf}% Remove "end if" text
\algtext*{EndFor}% Remove "end if" text
% ------------------
% -- Augment Data --
% ------------------
\begin{algorithm}

\caption{Expanded version of Pragmatic Feature Preference Augmentation}
\label{apend-alg:expanded_data_augment}
\vspace{1mm}
\textsc{Input:} $\mc{D}$ the preference dataset. \\
\textsc{Output:} $\mc{D}'$ the augmented preference dataset. \\

\begin{algorithmic}[1]
%\setstretch{1.2}
%
\State \textbf{Init:} $\mathcal{D'}=\mathcal{D}$
\For{$(a_1, a_2, f(a_1,a_2), \texttt{mask}) \in \mc{D}$}
\State $a_1^\circ, a_2^\circ = \texttt{mask}(a_1, a_2, f(a_1, a_2))$
\For{\text{active-feature-pairs} in \texttt{feat-combos}$(a_1^\circ, a_2^\circ)$}
\State $a_1' = \texttt{copy}(a_1)$
\State $a_2' = \texttt{copy}(a_2)$
    \For{$j_x, j_y$ in active-feature-pairs}
    \State $a_1'[j_x] = a_2[j_y]$
    \State $a_2'[j_y] = a_1[j_x]$
    \EndFor
    \State $f' = f(a_1, a_2)$
    \State $\mathcal{D}' = \mathcal{D}' \cup \{(a_1', a_2', f')\}$
\EndFor
\EndFor
\Return $\mc{D}'$
\end{algorithmic}
\end{algorithm}
\setlength{\textfloatsep}{15pt}
%%%%%%%%%%%%%%%%%%%%%%%%

% % for active_feat_pairs in feat_combos(a_1^circ, a_2^circ):
%     new_a_1 = copy(a_1)
%     new_a_2 = copy(a_2)
%     for idx_1, idx_2 in active_feat_pairs:
%         new_a_1[idx_1] = a_2[idx_2]
%         new_a_2[idx_2] = a_1[idx_1]
%     new_f = f(a_1, a_2)
%     $\mathcal{D}' = \mathcal{D}' \cup \{(new_a_1, new_a_2, new_f)\}$